\pdfoutput=1

\documentclass[11pt]{article}

\usepackage{acl}

\usepackage{times}
\usepackage{latexsym}
\usepackage[T1]{fontenc}
\usepackage[utf8]{inputenc}
\usepackage{microtype}
\usepackage{graphicx}
\usepackage{tabularx}
\usepackage{algorithm}
\usepackage{algpseudocode}
\usepackage{booktabs}
\usepackage{url}
\usepackage{xcolor}
\usepackage[framemethod=TikZ]{mdframed}
\definecolor{lightgray}{RGB}{245,245,245}

\title{Factored Verification: Detecting and Reducing \\ Hallucination in Summaries of Academic Papers}

\author{Charlie George \\
  Elicit \\
  \texttt{charlie@elicit.com} \\\And
  Andreas Stuhlm\"uller \\
  Elicit \\
  \texttt{andreas@elicit.com} \\}

\begin{document}
\maketitle
\begin{abstract}
Hallucination plagues even frontier LLMs---but how bad is it really for summarizing academic papers? We evaluate {\em Factored Verification}, a simple automated method for detecting hallucinations in abstractive summaries. This method sets a new SotA on hallucination detection in the summarization task of the HaluEval benchmark, achieving 76.2\% accuracy. We then use this method to estimate how often language models hallucinate when summarizing across multiple academic papers and find 0.62 hallucinations in the average ChatGPT (16k) summary, 0.84 for GPT-4, and 1.55 for Claude 2. We ask models to self-correct using {\em factored critiques} and find that this lowers the number of hallucinations to 0.49 for ChatGPT, 0.46 for GPT-4, and 0.95 for Claude 2. The hallucinations we find are often subtle, so we advise caution when using models to synthesize academic papers.
\end{abstract}

\section{Introduction}

Hallucination---the generation of inaccurate or ungrounded information---is a largely unsolved problem for LLMs \citep{kryscinski2019a, maynez2020, ji2023}. This is acceptable for creative use cases such as story generation and brainstorming, but would be highly problematic if common for academic summarization and Q\&A where factual accuracy is key. How common is hallucination for SotA models when answering questions given the abstracts of multiple scientific papers?

To answer this question, we first construct a simple method for checking hallucination inspired by \citet{kadavath2022} and \citet{lightman2023}: Given a summary, we automatically decompose it into key claims, assign a model-generated probability to each of the claims given the relevant sources, and combine these into an overall correctness probability. We validate this method on the hallucination detection benchmark HaluEval and set a new SotA, exceeding the previous chain-of-thought-based method by 10 absolute percentage points using the same language model.

\begin{figure}
  \vspace{-.3em}
  \centering
\includegraphics[width=0.5\textwidth]{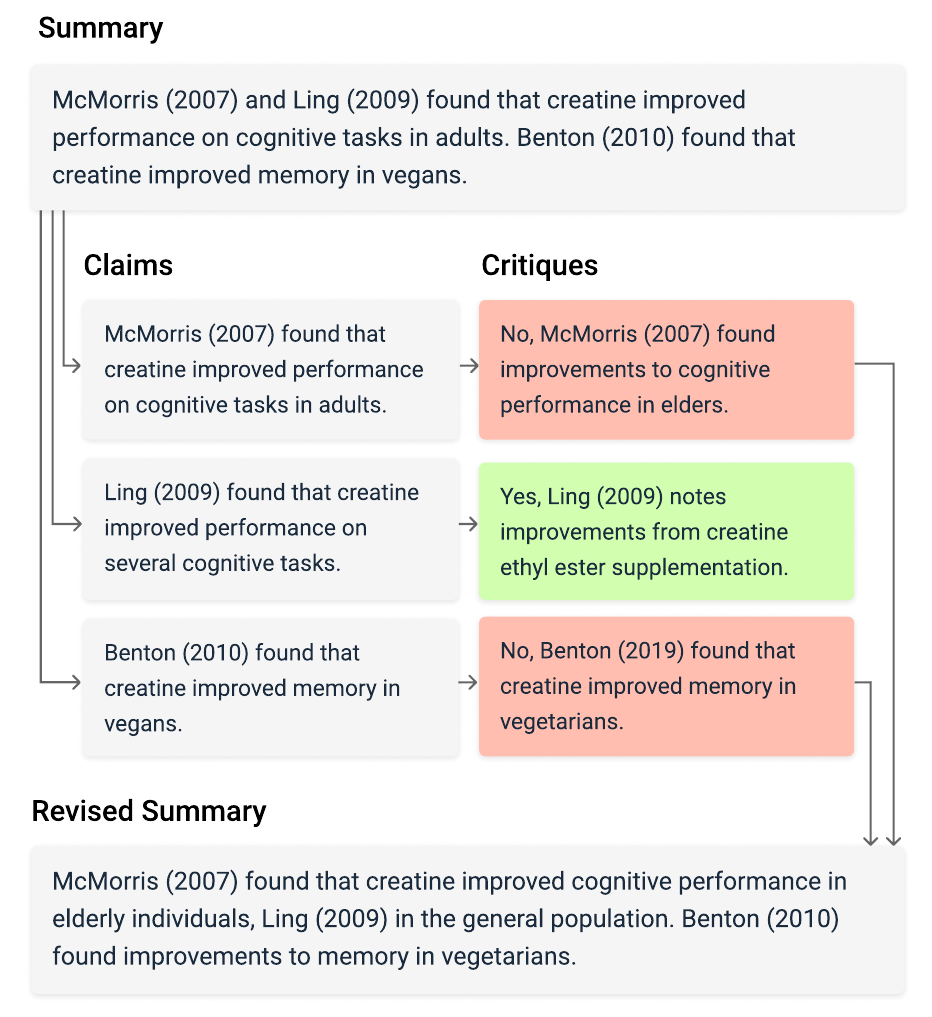}
  \caption{Factored Verification splits a summary into claims, checks each claim, and then optionally revises the summary to address the claim critiques. Each step is a language model task.}
  \label{fig:example_decomposition}
  \vspace{-.7em}
\end{figure}

We then apply Factored Verification to detecting hallucination in a real-world scientific summarization task. Given the abstracts of eight papers and a question, the task is to provide a question-relevant summary. We measure hallucination for SotA models including GPT-4 \citep{openai2023} and Claude 2 \citep{bai2022}, and estimate that the average summary has between 0.62 and 1.57 hallucinations.

Given that we can automatically detect some hallucinations, can we use this knowledge to reduce them? We treat the claim-wise critiques generated by Factored Verification as model-generated advice \citep{saunders2022a} and show that we can reduce detected hallucinations for every model we study, but that significant hallucination remains.


\section{Detecting hallucination with Factored Verification}

We first develop and validate Factored Verification, a simple method for using LLMs to detect hallucinations in settings where the relevant source material is provided.

\subsection{Defining ``hallucination''}

We call a claim ``hallucinated'' if it is not backed by the source material provided in context, even if it could be supported with other sources. For example, if the source material discusses the implementation of a public transport policy and the model-generated summary infers that the policy was aimed at addressing sustainability challenges, this is a hallucination unless the source explicitly talked about this as the goal of the policy.

\subsection{Method}
\label{sec:detection-method}

Following \citet{lightman2023}, we break each summary into a list of claims and then assign each claim a probability of being correct, both using LLM prompting. The claim decomposition prompt is in Appendix \ref{sec:fave-decompose}.

To compute the likelihood that a single claim is correct we use a few-shot prompt with GPT-4 base \cite{openai2023} and look up the probability of the final \texttt{Yes} token (Appendix \ref{sec:fave-verify}).
For ChatGPT, which doesn't provide access to token probabilities, we ask the model to verify that each claim is supported using few-shot chain-of-thought \cite{JasonChain}, interpreting the resulting Yes/No answer as a 0/1 probability (Appendix \ref{sec:fave-verify-chat}). 

Assuming independence of the correctness of claims for simplicity, the probability that the summary is correct is the product of the probabilities of each of the individual claims:

\begin{equation}
P_{\mathrm{summary}} = \prod_{i=1}^{n} P_{\mathrm{claim}_i}
\end{equation}

We classify a summary as hallucinated if $P_{\mathrm{summary}}$ is greater than a threshold $\theta$.

\subsection{Dataset}

To measure how well Factored Verification works, we use the summarization task of HaluEval, a hallucination benchmark \cite{li2023}. Each item in this task consists of a document and two summaries, one of which contains a subtle hallucination. 

Due to cost constraints, we randomly sample 5\% of the dataset as a ``training'' set, and 20\% as a test set (1000 and 4000 items respectively). The training set is only used to set the decision threshold $\theta$ by running GPT-4 with the aforementioned claim likelihood prompt on the training set examples and calculating the average of $P_{\mathrm{summary}}$ for both faithful and hallucinated summaries.

\subsection{Results}

Factored Verification is SotA for hallucination detection, exceeding prior few-shot and chain-of-thought approaches for both ChatGPT and GPT-4 (Table \ref{table:halueval}).

\begin{table}[h]
  \centering
  \begin{tabular}{c c c c}
    \toprule
    Model & Few-shot & CoT & Factored \\
    \midrule
    GPT-4 & 30.9\% & 75.5\% & \textbf{76.2\%} \\
    ChatGPT (3.5) & 58.5\%* & 61.2\%* & \textbf{71.2\%} \\
    \bottomrule
  \end{tabular}
  \caption{HaluEval summarization results, showing \% accuracy, $n = 4000$, * from \citet{li2023}.}
  \label{table:halueval}
\end{table}

For comparison, we also tested a single-prompt equivalent of Factored Verification, asking ChatGPT to write out each claim and verify it in one go. This achieved an accuracy of 63.3\%, which is below the multi-prompt factored method (71.2\%) and comparable to standard CoT results (61.2\%).

We were surprised by GPT-4's low $30.9\%$ accuracy when using the exact few-shot prompt from \citet{li2023}. To check our implementation, we ran the same code with ChatGPT and found an accuracy 59.9\% on our sample, roughly matching the 58.5\% from \citet{li2023}. Given that GPT-4 is much more competitive in the CoT and Factored Verification settings, we suspect that the \citet{li2023} few-shot prompt has idiosyncratic limitations.


\section{How often do models hallucinate in scientific summaries?}

We now turn to our task of interest, generating summaries based on the abstracts of scientific papers. We will apply Factored Verification to estimate how frequently models hallucinate in this context.

\subsection{Task}

Given a question and eight abstracts, the task is to generate a question-relevant summary where each claim is cited with the paper that backs it up. Figure \ref{fig:example-result} shows a simplified example with only two papers. The dataset of queries and abstracts\footnote{\url{https://github.com/elicit/fave-dataset}} is generated by using 200 anonymized search queries and their associated top results from Elicit \citep{elicit23}.

\begin{table*}
  \centering
  \begin{tabular}{p{3.5cm} p{3.5cm} p{3.5cm} p{3.5cm}}
    \toprule
    Model & Hallucinations per\newline summary (reported) & Hallucinations per \newline summary (adjusted) & \% of summaries with \newline reported hallucinations \\
    \midrule
    GPT-4 & $1.26 \rightarrow 0.69$ & $0.84 \rightarrow 0.46$ & $63.25\% \rightarrow 40\% $ \\
    ChatGPT (3.5, 16k) & $0.93 \rightarrow 0.735$ & $0.62 \rightarrow 0.49$ & $54\% \rightarrow 41.63\%$ \\
    Claude 2 & $2.32 \rightarrow 1.43$ & $1.55 \rightarrow 0.95$ & $83.0\% \rightarrow 71.50\%$ \\
    Claude instant & $2.35 \rightarrow 1.86$ & $1.57 \rightarrow 1.24$ & $87.0\% \rightarrow 81.50\%$\\
    \bottomrule
  \end{tabular}
  \caption{Prevalence of hallucination for models when generating summaries of academic papers, before and after revision with factored critiques. Based on manual inspection of approximately 100 data points our best guess is that the true prevalence of hallucination (``adjusted'') is 2/3 of the reports from automated evaluation.}
  \label{tab:results}
\end{table*}

\begin{figure}
    \centering
\begin{mdframed}[backgroundcolor=lightgray]
{\fontsize{9}{11}\selectfont

\textbf{Query:} What is the impact of creatine on cognition?

\textbf{Summary:} The papers suggest that creatine supplementation may have a positive impact on cognition. \textcolor{blue}{Ling (2009)} found that creatine ethyl ester supplementation improved performance on several cognitive tasks, although the results require replication using objective measures of compliance. \textcolor{blue}{Benton (2010)} found that creatine supplementation improved memory in vegetarians, but not in those who consume meat.
}
\end{mdframed}
    \caption{Example of the scientific summarization task, simplified from eight to two papers.}
    \label{fig:example-result}
\end{figure}

\subsection{Method}

We follow the hallucination detection strategy outlined in Section \ref{sec:detection-method}, with the following modifications to reduce compute cost:

\begin{enumerate}
  \item We directly treat sentences as claims.
  \item We only check claims that have associated citations.
  \item When checking each claim, we provide only the abstracts of the cited papers as sources.
\end{enumerate}

We expect that these modifications lead to little degradation given that almost all sentences have citations and the simplification step from sentence to claim is not doing much work.

\subsection{Results}

We run Factored Verification with ChatGPT, GPT-4, Claude 2, and Claude Instant \citep{openai2023, bai2022}. Table \ref{tab:results} shows that for all models, our method reports at least one hallucination in the majority of summaries.

We include additional results that show interactions between ChatGPT and GPT-4 when used as generation, criticism, and judge models in Table \ref{tab:interactions} in the Appendix. 

\subsection{Interpretation}

Based on the 76\% accuracy of Factored Verification on HaluEval, we know that there are likely false positives and/or false negatives, so we can't take the reported hallucination rates literally.

We manually inspected about a hundred claims evaluated by GPT-4. When GPT-4 said that a claim is supported, we agreed in all cases. When GPT-4 reported an unsupported claim, we agreed 66\% of the time. So, our best guess for the true hallucination rate is 2/3 of the reported hallucination rate.

Many of the claims we encountered were wrong in subtle ways that we would likely have missed without seeing the GPT-4 critiques, and would expect non-expert evaluators to miss, including:

\begin{itemize}
    \item Stating that a claim is supported by two abstracts when it is only supported by one
    \item Slightly exaggerating the findings of a paper
    \item Conflating the purpose of the study with the outcome
    \item Implying that two independent findings are linked
\end{itemize}

This augmentation of human evaluation is consistent with prior work by \citet{saunders_self-critiquing_2022} which found that model-generated critiques help humans find flaws in summaries.


\section{Reducing hallucination in scientific summaries with Factored Verification}

It is common for LLMs to apparently fail at a task, only to then succeed with better prompting. Can we prompt models using the detected inaccuracies to automatically reduce hallucination in scientific summaries?

\subsection{Baseline}

We ask GPT-4 to self-correct by first identifying false claims in its initial summary, then revising the summary given this correction (prompts in Appendix \ref{sec:appendix-self-correction-feedback} and \ref{sec:appendix-self-correction-revision}). This {\em increased} the average number of detected hallucinations from 1.55 to 2.13. \citet{huang_large_2023} similarly found that the GPT-4 generation of LLMs struggles to directly self-correct across a variety of reasoning datasets.

\subsection{Method}

To improve on the baseline, we propose to reduce hallucination with Factored Verification in three steps, as illustrated in Figure \ref{fig:example_decomposition} and shown in Algorithm \ref{alg:revision}. 

First, we create claim-wise critiques (true/false judgments and supporting reasoning) analogous to the hallucination detection method above: We ask the model to evaluate the supportedness of each sentence based on the cited abstracts. We then concatenate the critiques of the unsupported claims to form the {\em factored critique}. Finally we ask the model to revise the summary given that critique.

\begin{algorithm}
\caption{Factored Verification: Revising a summary by generating sentence-wise critiques}
\label{alg:revision}
\begin{algorithmic}[1]
\fontsize{9pt}{11pt}\selectfont
    \State Initialize empty list for critiques
    \For{each sentence in the summary}
        \State Critique $\gets$ LLM.critique(sentence, cited abstracts)
        \If{sentence is unsupported}
            \State Add Critique to the list of critiques
        \EndIf
    \EndFor
    \State FactoredCritique $\gets$ concat(critiques)
    \State RevisedSummary $\gets$ LLM.revise(FactoredCritique)
    \State \Return RevisedSummary
\end{algorithmic}
\end{algorithm}

Figure \ref{fig:example-critique} shows an example critique.

\begin{figure*}
\begin{mdframed}[backgroundcolor=lightgray]

{\fontsize{9}{11}\selectfont
\textbf{Critique:} \newline
\fontsize{9}{11}\selectfont- For the sentence {\em "\textcolor{blue}{Giuntella (2015)} found that immigration reduced waiting times for outpatient referrals, suggesting that policies affecting population dynamics can indirectly impact waiting times and thus healthcare costs."} in the summary: The claim accurately reflects the findings of \textcolor{blue}{Giuntella (2015)} that immigration reduced waiting times for outpatient referrals. However, the claim extends the findings of the study to suggest that policies affecting population dynamics can indirectly impact waiting times and thus healthcare costs. While this may be a reasonable inference, it is not directly supported by the abstract. \newline
\fontsize{9}{11}\selectfont- For the sentence {\em "\textcolor{blue}{Propper (2002)} found that GP fundholders were able to secure shorter waiting times for their patients, suggesting that giving family doctors the ability to choose the hospital where their patients are treated and the means to pay for some services can reduce waiting times."} in the summary: The claim is not directly supported by the abstract. The abstract states that the study investigates whether GP fundholders were able to secure shorter waiting times for their patients, but it does not provide the results of this investigation. Therefore, the claim that \textcolor{blue}{Propper (2002)} found that GP fundholders were able to secure shorter waiting times for their patients is not supported by the abstract.
}
\end{mdframed}
    \caption{Example of a factored critique generated with GPT-4}
    \label{fig:example-critique}
\end{figure*}

\subsection{Results}

Table \ref{tab:results} shows that Factord Verification reduces the number of summaries with reported hallucination by 5.50\% to 23.25\% (absolute) depending on the model, with ChatGPT being the lowest-hallucination model before critique (0.62 estimated hallucinations per summary), and GPT-4 being the lowest-hallucination model after critique and across all settings (0.46 estimated hallucinations per summary).

\section{Related work}

Hallucination is widely known to be a significant problem for LLMs \cite{luo2023a, peng2023, ji2023}, although to a much lesser extent for abstractive summarization where the information needed to answer is fully provided \cite{cao2022, huang2023}. 

Various strategies have been proposed to mitigate hallucination. Some strategies aim to prevent their occurrence by checking how familiar models are with instructions \cite{luo2023a}. Others, including our proposed method, focus on the detection and correction of hallucinations post-generation \cite{cao2022, huang2023a}.

Of these, notable strategies are the use of external knowledge and retrieval augmentation, and automated feedback \cite{shuster2021, peng2023, zhang2023}. While external knowledge is less relevant here, Factored Verification can be viewed as a kind of automated feedback.

In simultaneous work, \citet{dhuliawala2023} propose an automated feedback method called Chain-of-Verification, which is effectively the same as ours: (i) Draft an initial response, (ii) plan fact-checking questions, (iii) answer the questions independently, and (iv) generate a final revised response. Their evaluation focuses on out-of-context fact checking. Consistent with our results, they find that the factored version of their approach reduces hallucination for question-answering and text generation tasks.

\section{Relevance to AI alignment}

To align powerful AI systems, we need to be able to provide accurate feedback and supervision even when systems surpass human-level performance, a property known as scalable oversight \citep{amodei_concrete_2016}.

Today, reinforcement learning from human feedback (RLHF) is often used to align LLMs \cite{christiano_deep_2023, bai_training_2022, ouyang_training_2022}. However, GPT-4 already surpasses the performance of the average human on many academic tasks \cite{openai2023}, making it difficult for non-experts to provide effective reward signals. In our attempts to delegate evaluation of academic claims to non-expert contractors, we observed only 38\% inter-rater agreement for unsupported claims, a sign of similar difficulties. In the short term this can be solved by using contractors with specialized domain knowledge. However, this won't work if models surpass the capabilities of the best humans.

We have shown that factored critiques let models correct some of their own mistakes without need for human supervision. If similar approaches can be extended beyond hallucination reduction to richer tasks, they could help us scale supervision in lockstep with future model capabilities.

\section{Discussion}

Our main finding is that the absolute rate of hallucination of SotA models like ChatGPT, Claude 2, and GPT-4 is surprisingly high for academic summarization. This is true even with revision using factored critiques, which results in 0.46 to to 1.24 estimated hallucinations per summary.

A natural question to ask in this context is whether we can finetune on model-revised summaries, incrementally bootstraping to more and more accurate summaries, initially detecting and eliminating the most egregious failure modes, then more subtle ones with each training iteration.

Overall, despite incredible advances, language models still struggle with accurate summarization in academic contexts.  Many mistakes are only clear upon careful inspection of the sources and look identical to genuine answers otherwise. For now, we advise caution in situations where accuracy matters, as we would for human summaries as well.

\section{Acknowledgments}

We thank Noah Goodman, William Saunders, Jungwon Byun and James Brady for helpful early discussions. We would also like to thank Owain Evans and Justin Reppert for their feedback on early drafts. In addition, we are grateful to Ben Rachbach for help with human labeling and the rest of the Elicit team for supporting this work.

\newpage
\bibliography{factored-verification} 

\begin{thebibliography}{25}
\expandafter\ifx\csname natexlab\endcsname\relax\def\natexlab#1{#1}\fi

\bibitem[{Amodei et~al.(2016)Amodei, Olah, Steinhardt, Christiano, Schulman, and Mané}]{amodei_concrete_2016}
Dario Amodei, Chris Olah, Jacob Steinhardt, Paul Christiano, John Schulman, and Dan Mané. 2016.
\newblock \href {http://arxiv.org/abs/1606.06565} {Concrete {Problems} in {AI} {Safety}}.
\newblock ArXiv:1606.06565 [cs].

\bibitem[{Bai et~al.(2022{\natexlab{a}})Bai, Jones, Ndousse, Askell, Chen, DasSarma, Drain, Fort, Ganguli, Henighan, Joseph, Kadavath, Kernion, Conerly, {El-Showk}, Elhage, {Hatfield-Dodds}, Hernandez, Hume, Johnston, Kravec, Lovitt, Nanda, Olsson, Amodei, Brown, Clark, McCandlish, Olah, Mann, and Kaplan}]{bai2022}
Yuntao Bai, Andy Jones, Kamal Ndousse, Amanda Askell, Anna Chen, Nova DasSarma, Dawn Drain, Stanislav Fort, Deep Ganguli, Tom Henighan, Nicholas Joseph, Saurav Kadavath, Jackson Kernion, Tom Conerly, Sheer {El-Showk}, Nelson Elhage, Zac {Hatfield-Dodds}, Danny Hernandez, Tristan Hume, Scott Johnston, Shauna Kravec, Liane Lovitt, Neel Nanda, Catherine Olsson, Dario Amodei, Tom Brown, Jack Clark, Sam McCandlish, Chris Olah, Ben Mann, and Jared Kaplan. 2022{\natexlab{a}}.
\newblock \href {http://arxiv.org/abs/2204.05862} {Training a {{Helpful}} and {{Harmless Assistant}} with {{Reinforcement Learning}} from {{Human Feedback}}}.
\newblock \emph{arXiv:2204.05862 [cs]}.

\bibitem[{Bai et~al.(2022{\natexlab{b}})Bai, Jones, Ndousse, Askell, Chen, DasSarma, Drain, Fort, Ganguli, Henighan, Joseph, Kadavath, Kernion, Conerly, El-Showk, Elhage, Hatfield-Dodds, Hernandez, Hume, Johnston, Kravec, Lovitt, Nanda, Olsson, Amodei, Brown, Clark, McCandlish, Olah, Mann, and Kaplan}]{bai_training_2022}
Yuntao Bai, Andy Jones, Kamal Ndousse, Amanda Askell, Anna Chen, Nova DasSarma, Dawn Drain, Stanislav Fort, Deep Ganguli, Tom Henighan, Nicholas Joseph, Saurav Kadavath, Jackson Kernion, Tom Conerly, Sheer El-Showk, Nelson Elhage, Zac Hatfield-Dodds, Danny Hernandez, Tristan Hume, Scott Johnston, Shauna Kravec, Liane Lovitt, Neel Nanda, Catherine Olsson, Dario Amodei, Tom Brown, Jack Clark, Sam McCandlish, Chris Olah, Ben Mann, and Jared Kaplan. 2022{\natexlab{b}}.
\newblock \href {http://arxiv.org/abs/2204.05862} {Training a {Helpful} and {Harmless} {Assistant} with {Reinforcement} {Learning} from {Human} {Feedback}}.
\newblock ArXiv:2204.05862 [cs].

\bibitem[{Cao et~al.(2022)Cao, Dong, and Cheung}]{cao2022}
Meng Cao, Yue Dong, and Jackie Cheung. 2022.
\newblock \href {https://doi.org/10.18653/v1/2022.acl-long.236} {Hallucinated but {{Factual}}! {{Inspecting}} the {{Factuality}} of {{Hallucinations}} in {{Abstractive Summarization}}}.
\newblock In \emph{Proceedings of the 60th {{Annual Meeting}} of the {{Association}} for {{Computational Linguistics}} ({{Volume}} 1: {{Long Papers}})}, pages 3340--3354, {Dublin, Ireland}. {Association for Computational Linguistics}.

\bibitem[{Christiano et~al.(2023)Christiano, Leike, Brown, Martic, Legg, and Amodei}]{christiano_deep_2023}
Paul Christiano, Jan Leike, Tom~B. Brown, Miljan Martic, Shane Legg, and Dario Amodei. 2023.
\newblock \href {http://arxiv.org/abs/1706.03741} {Deep reinforcement learning from human preferences}.
\newblock ArXiv:1706.03741 [cs, stat].

\bibitem[{Dhuliawala et~al.(2023)Dhuliawala, Komeili, Xu, Raileanu, Li, Celikyilmaz, and Weston}]{dhuliawala2023}
Shehzaad Dhuliawala, Mojtaba Komeili, Jing Xu, Roberta Raileanu, Xian Li, Asli Celikyilmaz, and Jason Weston. 2023.
\newblock \href {https://doi.org/10.48550/arXiv.2309.11495} {Chain-of-{{Verification Reduces Hallucination}} in {{Large Language Models}}}.

\bibitem[{Huang et~al.(2023{\natexlab{a}})Huang, Chen, Mishra, Zheng, Yu, Song, and Zhou}]{huang_large_2023}
Jie Huang, Xinyun Chen, Swaroop Mishra, Huaixiu~Steven Zheng, Adams~Wei Yu, Xinying Song, and Denny Zhou. 2023{\natexlab{a}}.
\newblock \href {http://arxiv.org/abs/2310.01798} {Large {Language} {Models} {Cannot} {Self}-{Correct} {Reasoning} {Yet}}.
\newblock ArXiv:2310.01798 [cs].

\bibitem[{Huang et~al.(2023{\natexlab{b}})Huang, Chan, and Ji}]{huang2023a}
Kung-Hsiang Huang, Hou~Pong Chan, and Heng Ji. 2023{\natexlab{b}}.
\newblock \href {https://doi.org/10.18653/v1/2023.acl-long.311} {Zero-shot {{Faithful Factual Error Correction}}}.
\newblock In \emph{Proceedings of the 61st {{Annual Meeting}} of the {{Association}} for {{Computational Linguistics}} ({{Volume}} 1: {{Long Papers}})}, pages 5660--5676, {Toronto, Canada}. {Association for Computational Linguistics}.

\bibitem[{Huang et~al.(2023{\natexlab{c}})Huang, Feng, Feng, and Qin}]{huang2023}
Yichong Huang, Xiachong Feng, Xiaocheng Feng, and Bing Qin. 2023{\natexlab{c}}.
\newblock \href {https://doi.org/10.48550/arXiv.2104.14839} {The {{Factual Inconsistency Problem}} in {{Abstractive Text Summarization}}: {{A Survey}}}.

\bibitem[{{Jason Wei} et~al.(){Jason Wei}, {Xuezhi Wang}, {Dale Schuurmans}, {Maarten Bosma}, {Ed Chi}, {Quoc Le}, and {Denny Zhou}}]{JasonChain}
{Jason Wei}, {Xuezhi Wang}, {Dale Schuurmans}, {Maarten Bosma}, {Ed Chi}, {Quoc Le}, and {Denny Zhou}.
\newblock Chain of {{Thought Prompting Elicits Reasoning}} in {{Large Language Models}}.

\bibitem[{Ji et~al.(2023)Ji, Lee, Frieske, Yu, Su, Xu, Ishii, Bang, Dai, Madotto, and Fung}]{ji2023}
Ziwei Ji, Nayeon Lee, Rita Frieske, Tiezheng Yu, Dan Su, Yan Xu, Etsuko Ishii, Yejin Bang, Wenliang Dai, Andrea Madotto, and Pascale Fung. 2023.
\newblock \href {https://doi.org/10.1145/3571730} {Survey of {{Hallucination}} in {{Natural Language Generation}}}.
\newblock \emph{ACM Computing Surveys}, 55(12):1--38.

\bibitem[{Kadavath et~al.(2022)Kadavath, Conerly, Askell, Henighan, Drain, Perez, Schiefer, {Hatfield-Dodds}, DasSarma, {Tran-Johnson}, Johnston, {El-Showk}, Jones, Elhage, Hume, Chen, Bai, Bowman, Fort, Ganguli, Hernandez, Jacobson, Kernion, Kravec, Lovitt, Ndousse, Olsson, Ringer, Amodei, Brown, Clark, Joseph, Mann, McCandlish, Olah, and Kaplan}]{kadavath2022}
Saurav Kadavath, Tom Conerly, Amanda Askell, Tom Henighan, Dawn Drain, Ethan Perez, Nicholas Schiefer, Zac {Hatfield-Dodds}, Nova DasSarma, Eli {Tran-Johnson}, Scott Johnston, Sheer {El-Showk}, Andy Jones, Nelson Elhage, Tristan Hume, Anna Chen, Yuntao Bai, Sam Bowman, Stanislav Fort, Deep Ganguli, Danny Hernandez, Josh Jacobson, Jackson Kernion, Shauna Kravec, Liane Lovitt, Kamal Ndousse, Catherine Olsson, Sam Ringer, Dario Amodei, Tom Brown, Jack Clark, Nicholas Joseph, Ben Mann, Sam McCandlish, Chris Olah, and Jared Kaplan. 2022.
\newblock \href {https://doi.org/10.48550/arXiv.2207.05221} {Language {{Models}} ({{Mostly}}) {{Know What They Know}}}.

\bibitem[{Kry{\'s}ci{\'n}ski et~al.(2019)Kry{\'s}ci{\'n}ski, McCann, Xiong, and Socher}]{kryscinski2019a}
Wojciech Kry{\'s}ci{\'n}ski, Bryan McCann, Caiming Xiong, and Richard Socher. 2019.
\newblock \href {https://doi.org/10.48550/arXiv.1910.12840} {Evaluating the {{Factual Consistency}} of {{Abstractive Text Summarization}}}.

\bibitem[{Li et~al.(2023)Li, Cheng, Zhao, Nie, and Wen}]{li2023}
Junyi Li, Xiaoxue Cheng, Wayne~Xin Zhao, Jian-Yun Nie, and Ji-Rong Wen. 2023.
\newblock \href {https://doi.org/10.48550/arXiv.2305.11747} {{{HaluEval}}: {{A Large-Scale Hallucination Evaluation Benchmark}} for {{Large Language Models}}}.

\bibitem[{Lightman et~al.(2023)Lightman, Kosaraju, Burda, Edwards, Baker, Lee, Leike, Schulman, Cobbe, and Sutskever}]{lightman2023}
Hunter Lightman, Vineet Kosaraju, Yura Burda, Harri Edwards, Bowen Baker, Teddy Lee, Jan Leike, John Schulman, Karl Cobbe, and Ilya Sutskever. 2023.
\newblock Let's {{Verify Step}} by {{Step}}.

\bibitem[{Luo et~al.(2023)Luo, Xiao, and Ma}]{luo2023a}
Junyu Luo, Cao Xiao, and Fenglong Ma. 2023.
\newblock \href {https://doi.org/10.48550/arXiv.2309.02654} {Zero-{{Resource Hallucination Prevention}} for {{Large Language Models}}}.

\bibitem[{Maynez et~al.(2020)Maynez, Narayan, Bohnet, and McDonald}]{maynez2020}
Joshua Maynez, Shashi Narayan, Bernd Bohnet, and Ryan McDonald. 2020.
\newblock \href {https://doi.org/10.18653/v1/2020.acl-main.173} {On {{Faithfulness}} and {{Factuality}} in {{Abstractive Summarization}}}.
\newblock In \emph{Proceedings of the 58th {{Annual Meeting}} of the {{Association}} for {{Computational Linguistics}}}, pages 1906--1919, {Online}. {Association for Computational Linguistics}.

\bibitem[{OpenAI(2023)}]{openai2023}
OpenAI. 2023.
\newblock \href {http://arxiv.org/abs/2303.08774} {{{GPT-4 Technical Report}}}.

\bibitem[{Ouyang et~al.(2022)Ouyang, Wu, Jiang, Almeida, Wainwright, Mishkin, Zhang, Agarwal, Slama, Ray, Schulman, Hilton, Kelton, Miller, Simens, Askell, Welinder, Christiano, Leike, and Lowe}]{ouyang_training_2022}
Long Ouyang, Jeff Wu, Xu~Jiang, Diogo Almeida, Carroll~L. Wainwright, Pamela Mishkin, Chong Zhang, Sandhini Agarwal, Katarina Slama, Alex Ray, John Schulman, Jacob Hilton, Fraser Kelton, Luke Miller, Maddie Simens, Amanda Askell, Peter Welinder, Paul Christiano, Jan Leike, and Ryan Lowe. 2022.
\newblock \href {http://arxiv.org/abs/2203.02155} {Training language models to follow instructions with human feedback}.
\newblock ArXiv:2203.02155 [cs].

\bibitem[{Peng et~al.(2023)Peng, Galley, He, Cheng, Xie, Hu, Huang, Liden, Yu, Chen, and Gao}]{peng2023}
Baolin Peng, Michel Galley, Pengcheng He, Hao Cheng, Yujia Xie, Yu~Hu, Qiuyuan Huang, Lars Liden, Zhou Yu, Weizhu Chen, and Jianfeng Gao. 2023.
\newblock \href {https://doi.org/10.48550/arXiv.2302.12813} {Check {{Your Facts}} and {{Try Again}}: {{Improving Large Language Models}} with {{External Knowledge}} and {{Automated Feedback}}}.

\bibitem[{Saunders et~al.(2022{\natexlab{a}})Saunders, Yeh, Wu, Bills, Ouyang, Ward, and Leike}]{saunders2022a}
William Saunders, Catherine Yeh, Jeff Wu, Steven Bills, Long Ouyang, Jonathan Ward, and Jan Leike. 2022{\natexlab{a}}.
\newblock Self-critiquing models for assisting human evaluators.
\newblock https://arxiv.org/abs/2206.05802v2.

\bibitem[{Saunders et~al.(2022{\natexlab{b}})Saunders, Yeh, Wu, Bills, Ouyang, Ward, and Leike}]{saunders_self-critiquing_2022}
William Saunders, Catherine Yeh, Jeff Wu, Steven Bills, Long Ouyang, Jonathan Ward, and Jan Leike. 2022{\natexlab{b}}.
\newblock \href {https://doi.org/10.48550/arXiv.2206.05802} {Self-critiquing models for assisting human evaluators}.
\newblock ArXiv:2206.05802 [cs].

\bibitem[{Shuster et~al.(2021)Shuster, Poff, Chen, Kiela, and Weston}]{shuster2021}
Kurt Shuster, Spencer Poff, Moya Chen, Douwe Kiela, and Jason Weston. 2021.
\newblock \href {https://doi.org/10.48550/arXiv.2104.07567} {Retrieval {{Augmentation Reduces Hallucination}} in {{Conversation}}}.

\bibitem[{Stuhlm{\"u}ller and Byun(2023)}]{elicit23}
Andreas Stuhlm{\"u}ller and Jungwon Byun. 2023.
\newblock Elicit: {{The AI Research Assistant}}.
\newblock https://elicit.org/.

\bibitem[{Zhang et~al.(2023)Zhang, Pan, Zhao, and Wang}]{zhang2023}
Shuo Zhang, Liangming Pan, Junzhou Zhao, and William~Yang Wang. 2023.
\newblock \href {https://doi.org/10.48550/arXiv.2305.13669} {Mitigating {{Language Model Hallucination}} with {{Interactive Question-Knowledge Alignment}}}.

\end{thebibliography}
\bibliographystyle{acl_natbib}

\newpage
\appendix

\section{Appendix}
\label{sec:appendix}

\subsection{Factored Verification prompts (HaluEval)}

\subsubsection{Decomposing a summary into claims}
\label{sec:fave-decompose}

{\small \texttt{Below is a summary of a document. Please extract ALL the claims from the document. You should give your answer as a list separated by "-" and start by saying "The claims are:"\newline\newline[summary].}}

\subsubsection{Verifying the correctness of a claim with GPT-4 base}
\label{sec:fave-verify}

{\small \texttt{Below are a set of documents and claims. We will check if the document that the claim is supported by the document or otherwise inaccurate. Below are some examples. It can sometimes be the case that a claim is very subtly wrong.\newline\newline [Few-shot examples]\newline\newline Example 5:\newline\newline Document: [Document]\newline\newline Claim: [Claim]\newline\newline Supported: Yes}}

\subsubsection{Verifying the correctness of a claim with ChatGPT}
\label{sec:fave-verify-chat}

{\small \texttt{Below is a claim and a document. Check that the claim is supported by the document. If it is, say "Yes". If it is not, say "No". \newline\newline Document: [document]\newline\newline Claim: [claim]\newline\newline Give your answer in the following format:\newline\newline Reasoning: [give your reasoning (including quotes) here]\newline Supported: [Yes/No]\newline\newline Remember you MUST include quotes in your reasoning.}}

\subsection{Prompt template for generating summaries of academic papers}

{\small \texttt{I now need you to help me summarize many more papers in the same way as above. Our research question is "[question]".\newline\newline I've collected many papers that might address this research question.\newline\newline Paper [number]: [reference]\newline\newline Title: [title]\newline\newline Abstract: [abstract]\newline\newline Write a summary of what the papers collectively say about the research question. Use the same format as the summary above.\newline\newline You must cite the papers in your summary. You can use the following format: Author (year)\newline\newline You will only include the findings that directly answer our research question, ignoring other findings that are only loosely relevant. Remember to include citations in the final summary. Your final summary should use varied and engaging language.}}

\subsection{Prompt templates for Factored Verification (academic papers)}

\subsubsection{Generating claim-wise critiques}

{\small \texttt{I need some more help verifying some claims from scientific papers.\newline\newline The claim is from [paper references]:\newline\newline [reference]:\newline\newline Title: [title]\newline\newline Abstract: [abstract]\newline\newline ==\newline\newline Claim: [claim]\newline\newline First give a critique of the claim. Then, say whether it is supported by the abstract["s" if we have multiple abstracts]. Finally, if claim is not supported give a revised claim that is supported by the abstract["s" if we have multiple abstracts].\newline\newline If the claim is partially supported say "No" for the "Supported" field and give a revised claim that is fully supported by the abstract.\newline\newline Format:\newline\newline Critique: [critique]\newline Supported: "Yes" or "No"\newline Revised Claim: [revised claim] or "N/A" if claim is supported.}}

\subsubsection{Revision based on claim-wise critiques}

As a follow-up to the papers and model-provided summary:

{\small \texttt{Ok, after reading your summary, I have some feedback:\newline\newline Feedback: \newline\newline I have some concerns about the factual accuracy of the summary:\newline\newline - For the sentence "[original false claim]" in the summary: [critique]\newline\newline ===\newline\newline Can you correct your summary incorporating each piece of my feedback? The concerns are MOST important to address. Start by writing "Corrected summary:" and then your corrected summary. Keep everything not mentioned in my feedback the same.}}

\subsection{Prompt templates for self-correction baseline (academic papers)}

\subsubsection{Generating self-correction feedback}
\label{sec:appendix-self-correction-feedback}

{\small \texttt{Below is a list of academic papers.\newline\newline [Papers]\newline\newline This is a summary of the papers:\newline\newline [summary]\newline\newline Please read the papers and the summary and give feedback. The feedback should ONLY look the at factual accuracy of the summary and make sure that any claims made are FULLY supported by the relevant papers. Write "Feedback:" and then your feedback. You should give a VERY harsh long and detailed piece of feedback.}}

\subsection{Revision based on self-generated feedback}
\label{sec:appendix-self-correction-revision}

{\small \texttt{Ok, after reading your summary, I have some feedback:\newline\newline Feedback: \newline\newline [Model feedback from prompt above]\newline\newline Can you correct your summary incorporating each piece of my feedback? The concerns are MOST important to address. Start by writing "Corrected summary:" and then your corrected summary. Keep everything not mentioned in my feedback the same.}}

\subsection{Additional results}

See Table \ref{tab:interactions}.

\begin{table*}[h]
\centering
\begin{tabular}{p{2.1cm}p{2.1cm}p{2.1cm}p{3.2cm}p{3.5cm}}
\toprule
Summary \newline model & Critique \newline model & Judge & Hallucinations per \newline summary (reported) & \% of summaries with \newline reported hallucinations \\ \midrule
ChatGPT & - & GPT-4 & 0.89 & 51.00\% \\
ChatGPT & ChatGPT & GPT-4 & 0.98 & 52.00\% \\
ChatGPT & GPT-4 & GPT-4 & 0.45 & 28.00\% \\
GPT-4 & - & GPT-4 & 1.55 & 69.50\% \\
GPT-4 & ChatGPT & GPT-4 & 1.19 & 67.00\% \\
GPT-4 & GPT-4 & GPT-4 & 0.51 & 29.50\% \\
GPT-4 & - & ChatGPT & 0.84 & 48.00\% \\
GPT-4 & ChatGPT & ChatGPT & 0.37 & 23.50\% \\
ChatGPT & - & ChatGPT & 0.97 & 57.00\% \\
ChatGPT & GPT-4 & ChatGPT & 0.85 & 49.50\% \\
ChatGPT & ChatGPT & ChatGPT & 0.66 & 37.00\% \\ \bottomrule
\end{tabular}
\caption{Interaction effects between ChatGPT and GPT-4 as summary, critique, and judge models. ChatGPT refers to the GPT-3.5 series with 16k context. Dashes indicate that no revision was used. In the main paper, for ChatGPT and GPT-4, we average over \{ ChatGPT, GPT-4 \} as critique generation and evaluation models to reduce interaction effects. For Claude models, we use ChatGPT as a judge.}
\label{tab:interactions}
\end{table*}

\end{document}